\documentclass[letterpaper, 10 pt, journal, twoside]{IEEEtran}

\usepackage[backend=biber,
            hyperref=true,
            url=false,
            isbn=false,
            doi=false,
            backref=false,
            style=ieee,
            citestyle=numeric-comp,
            sorting=nyt,
            block=none]{biblatex}

\usepackage[utf8]{inputenc}

\usepackage{float}
\usepackage{booktabs}
\usepackage{graphicx}
\usepackage{hyperref}
\usepackage{multirow}
\usepackage{balance}
\usepackage{placeins}
\usepackage{soul}
\usepackage{standalone}
\usepackage[font={footnotesize}]{caption}
\usepackage{subcaption}
\usepackage{tikz}
\usepackage{url}
\usepackage{amsmath, amssymb}
\usepackage{dsfont}
\usepackage[ruled]{algorithm2e}
\usepackage{algorithmic}
\usepackage{wrapfig}
\usepackage{soul}
\usepackage{times}
\usepackage{latexsym}
\usepackage{amsfonts}
\usepackage{booktabs}
\usepackage{bbm}
\usepackage{multicol}
\usepackage{hyperref}
\usepackage{caption}
\usepackage{subcaption}
\usepackage[normalem]{ulem}
\usepackage[T1]{fontenc}
\newcommand{\name}[1]{PRIME}

\title{\scalebox{0.93}{PRIME: Scaffolding Manipulation Tasks with Behavior}\\ \scalebox{0.93}{Primitives for Data-Efficient Imitation Learning}}

\author{Tian Gao$^{1*}$, Soroush Nasiriany$^{2}$, Huihan Liu$^{2}$, Quantao Yang$^{3*}$, Yuke Zhu$^{2}$
\thanks{Manuscript received: March 1, 2024; Revised May 29, 2024; Accepted July 1, 2024.}
\thanks{This paper was recommended for publication by Editor Aleksandra Faust upon evaluation of the Associate Editor and Reviewers' comments.
}
\thanks{$^{1}$ Tian Gao is with the Department of Computer Science, Stanford University, {\tt\footnotesize tiangao@stanford.edu}}
\thanks{$^{2}$Soroush Nasiriany, Huihan Liu, and Yuke Zhu are with the Department of Computer Science, the University of Texas at Austin.}
\thanks{$^{3}$ Quantao Yang is with the Department of Computer Science, KTH Royal Institute of Technology. }
\thanks{$^{*}$ This work was done when Tian Gao and Quantao Yang were visiting researchers at UT Austin.}%
\thanks{Digital Object Identifier (DOI): see top of this page.}
}

\begin{document}

\maketitle
    
\begin{abstract}
Imitation learning has shown great potential for enabling robots to acquire complex manipulation behaviors. However, these algorithms suffer from high sample complexity in long-horizon tasks, where compounding errors accumulate over the task horizons. We present \name{} (\underline{PR}imitive-based \underline{IM}itation with data \underline{E}fficiency), a behavior primitive-based framework designed for improving the data efficiency of imitation learning.
\name{} scaffolds robot tasks by decomposing task demonstrations into primitive sequences, followed by learning a high-level control policy to sequence primitives through imitation learning.
Our experiments demonstrate that \name{} achieves a significant performance improvement in multi-stage manipulation tasks, with 10-34\% higher success rates in simulation over state-of-the-art baselines and 20-48\% on physical hardware.\footnote{Additional materials are available at \url{https://ut-austin-rpl.github.io/PRIME/}.}
\end{abstract}


\section{Introduction}
\IEEEPARstart{I}{mitation} learning (IL) has become a powerful paradigm for programming robots to perform manipulation tasks.
Policies trained through imitation have exhibited diverse and complex behaviors, such as assembling parts~\cite{zhao2023aloha}, preparing coffee~\cite{zhu2022viola}, making pizza~\cite{chi2023diffusionpolicy}, and folding cloth~\cite{wang2023mimicplay}. 
Deep IL methods aim at training policies that map sensory observations directly to low-level motor commands~\cite{rt12022,mandlekar2021matters,chi2023diffusionpolicy}. While conceptually simple, these methods usually require a large volume of human demonstrations, making them costly for tackling long-horizon tasks. Furthermore, the direct imitation of low-level motor actions leads to limited generalization abilities of the learned policy.

One solution to improve data efficiency and model generalization is incorporating temporal abstraction into policy learning~\cite{precup2000temporal}.
Conventional methods afforded with temporal abstraction decouple learning new tasks into learning \textit{what} subtasks to perform and \textit{how} to achieve them. Among these methods, \textit{skills} represent a popular form of temporal abstraction, offering a systematic approach to decomposing complex tasks for robots. Skills serve as fundamental building blocks, capturing necessary robot behaviors for specific tasks, such as grasping an object.
The first step of incorporating skills as temporal abstraction is to obtain a repertoire of motor skills that capture \textit{how} to perform behaviors, serving as reusable building blocks for various tasks.
This reduces the problem of learning new tasks to learning \textit{what} behaviors to perform rather than \textit{how} to perform them, simplifying the learning process and enhancing generalization. The second step involves learning a policy for skill sequencing.
To acquire skills, one popular approach is skill learning, which learns low-level skills that capture short-horizon sequences of robot actions by learning either continuous latent skill representations  \cite{nasiriany2022learning,ajay2020opal,pertsch2021accelerating} or a discrete set of skills with continuous parameters ~\cite{shankar2020discovering, zhu2022bottom}. 
Prior work in skill learning extracts skills from a large amount of prior human data. 
While promising, a core limitation of these methods is the need for substantial human data to ensure the learned skills possess a high generalization capability.

\begin{figure}
  \includegraphics[width=\linewidth]{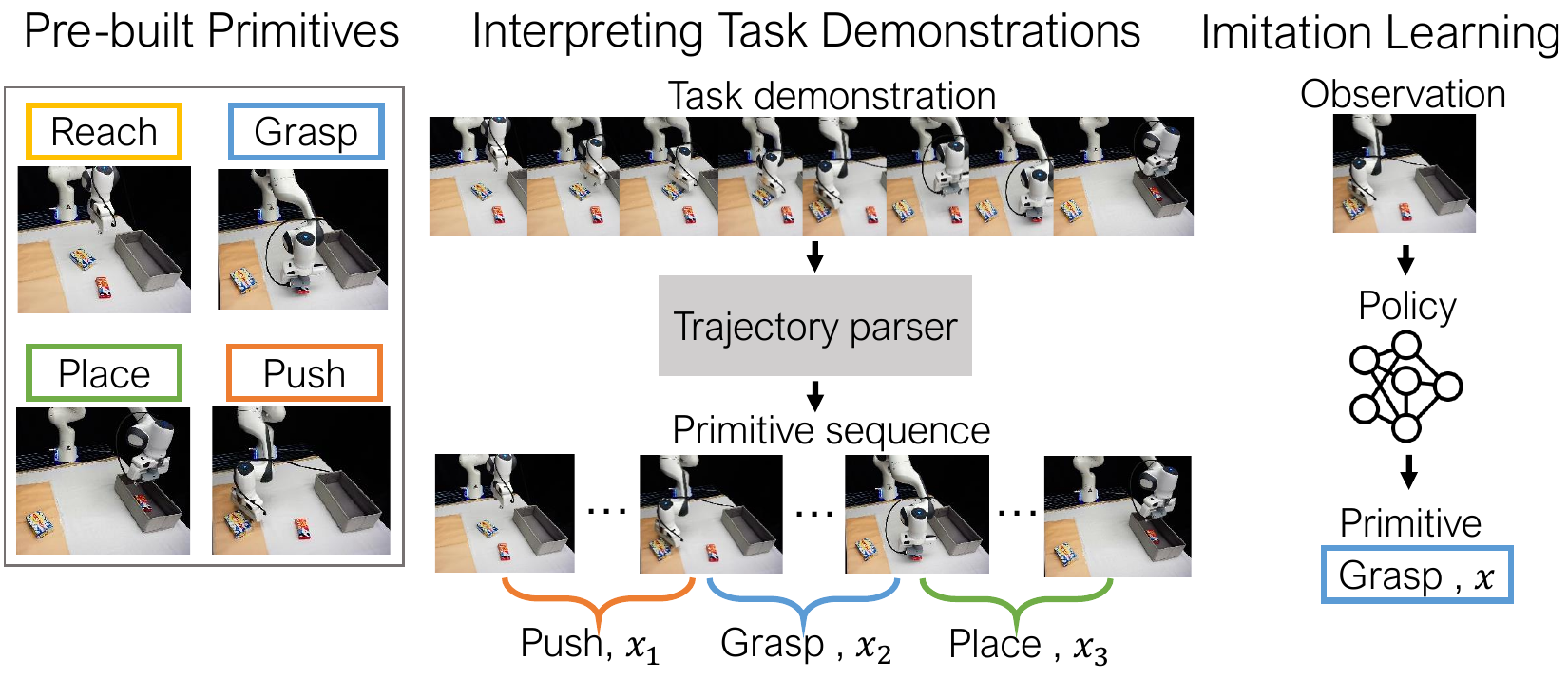}
  \caption{\textbf{Overview of PRIME.} (Left) Our learning framework leverages a set of pre-built behavior primitives to scaffold manipulation tasks. (Middle) Given task demonstrations, we use a trajectory parser to parse each demonstration into a sequence of primitive types (such as ``push'', ``grasp'' and ``place'') and their corresponding parameters $x_i$. (Right) With these parsed sequences of primitives, we use imitation learning to acquire a policy capable of predicting primitive types (such as ``grasp'') and corresponding parameters $x$ based on observations.
  } 
  \label{fig:pull}
  \vspace{1mm}
\end{figure}

\begin{figure*}
    \centering
      \includegraphics[width=0.9\linewidth]{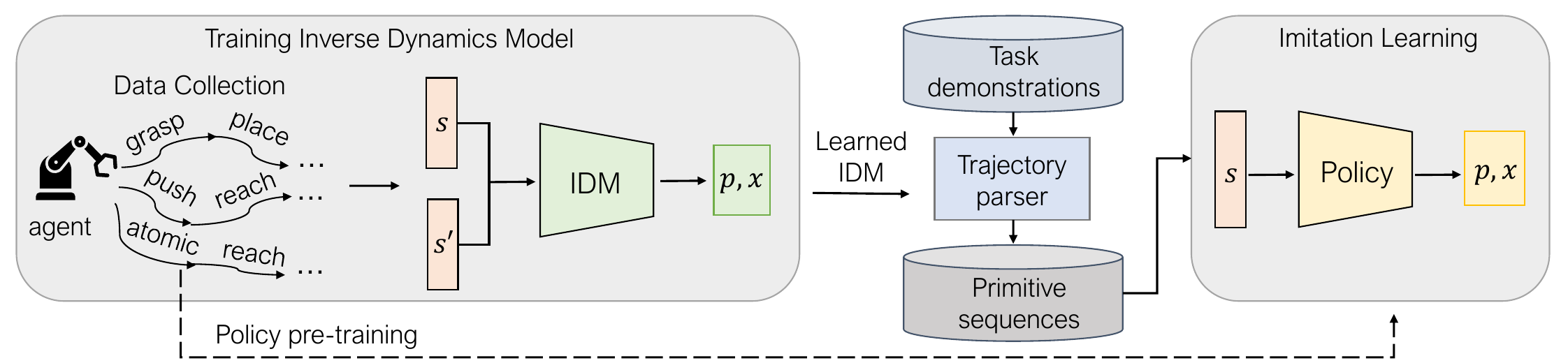}
      \caption{\textbf{Method Overview.} We develop a self-supervised data collection procedure that randomly executes sequences of behavior primitives in the environment. With the generated dataset, we train an IDM that maps an initial state $s$ and a final state $s'$ from segments in task demonstrations to a primitive type $p$ and corresponding parameters $x$. To derive the optimal primitive sequences, we build a trajectory parser capable of parsing task demonstrations into primitive sequences using the learned IDM. Finally, we train the policy using parsed primitive sequences.}
      \label{fig:pipeline}
    \vspace{2mm}
    \end{figure*}
    
Recent work has explored using robotic \textit{behavior primitives}~\cite{nasiriany2022augmenting, dalal2021accelerating,chen2023predicting,luo2023multi} to decompose manipulation tasks, such as movement primitives~\cite{ijspeert2013dmps, neumann2014learning}, motion planning~\cite{toussaint2015logic, garrett2021integrated, lozano2014constraint}, and grasping systems~\cite{mahler2017dexnet2, bohg2013data}.
A behavior primitive is a parameterized module designed to capture a certain movement pattern, usually with explicit semantic meaning (\textit{e.g.}, grasping). \textcolor{black}{The input parameters instantiate the behavior primitive into a specific movement, with the output being a sequence of motor actions to control the robot.} 
These primitives enjoy the advantages of re-usability, modularity, and robustness toward variations.
To utilize these primitives, recent work has proposed learning high-level policies that discover the optimal sequence of primitives using reinforcement learning (RL)~\cite{nasiriany2022augmenting, chitnis2020efficient, dalal2021accelerating}.
However, RL requires expensive exploration even in the action space afforded by the primitives and is unsafe to train on real robots.
Another notable line of work learns the policy with primitives from segmented demonstrations using imitation learning~\cite{shiarlis2018taco, huang2019neural, mahmoudieh2020weakly}. Segmented demonstrations require costly human efforts to manually segment demonstrations into primitive sequences.

In response to the above challenges, we introduce PRIME (\underline{PR}imitive-based \underline{IM}itation with data \underline{E}fficiency), a data-efficient imitation learning framework based on behavior primitives (see Fig. \ref{fig:pull}). We provide a small set of task demonstrations with raw sensory observations and a discrete collection of behavior primitives. Our framework consists of a two-step learning process: first, parsing task demonstrations as primitive sequences via a trajectory parser without the need for any human annotations, and subsequently, training a policy through imitation learning to predict the sequence of primitives (such as ``push'' and ``grasp'') and their corresponding parameters given observations.
By incorporating primitives, we break down long-horizon tasks into shorter sequences of primitives, significantly reducing the complexity and temporal horizon for imitation learning.

In this work, PRIME does not require access to segmentation labels, rendering the parsing of demonstrations a challenging task. To segment demonstrations into primitive sequences, it is essential to establish a mapping from raw observation-action sequences to primitives.
To generate the data necessary for learning the mapping, we introduce a self-supervised data collection procedure~\cite{pinto2016supersizing,chen2023predicting} that randomly samples sequences of primitives to execute within the environment, effectively reducing the need for human efforts in collecting prior data. 
Subsequently, we train an Inverse Dynamics Model (IDM)~\cite{paster2020planning, zheng2023semi, brandfonbrener2023inverse, pavse2020ridm, du2024learning} on the collected data, which maps pairs of states to primitives. 
We use this IDM with a dynamic programming algorithm to identify the optimal primitive sequences derived from task demonstrations. 

We evaluate our method's effectiveness in tabletop manipulation tasks, both in simulation and on real hardware. Our results highlight PRIME's substantial performance gains over state-of-the-art imitation learning baselines in a low-data regime. In simulations, success rates increase by 10.0\% to 33.6\%, and on real robots, by 20.0\% to 48.3\%. We further verify that our trajectory parser can effectively parse task demonstrations into primitive sequences that can be replayed to accomplish the task with success rates exceeding 90\%. Moreover, our IDM generalizes to unseen environments, achieving performance levels comparable to those in training environments.

We highlight three contributions of this work: 1) We introduce PRIME, a data-efficient imitation learning framework that scaffolds robot tasks with behavior primitives; 2) We develop a trajectory parser that transforms task demonstrations into primitive sequences using dynamic programming without segmentation labels; 3) We validate the effectiveness of PRIME in simulation and on real hardware. 
\section{Related Work} 
\subsection{Learning from Demonstration}
Learning from Demonstration (LfD) has shown promise in robot manipulation tasks~\cite{ravichandar2020recent, chernova_gmm, paraschos2013probabilistic, traPPCA}. LfD aims to enable an agent to observe and replicate expert behavior to effectively achieve a designated task. Within the domain of LfD, a diverse and extensive range of approaches has emerged, encompassing imitation learning~\cite{mandlekar2021matters, hussein2017imitation, mandlekar2020learning}, demonstration-guided RL~\cite{pertsch2021accelerating, pertsch2021guided, singh2020parrot}, and offline RL~\cite{ajay2020opal, mandlekar2020iris, kumar2022pre, yu2022uds}. These approaches that rely on demonstration guidance often necessitate a substantial number of expert demonstrations, thereby limiting their data efficiency.
To reduce the burdens of collecting expert demonstrations, a common line of work learns from task-agnostic play data~\cite{lynch2020learning, mees2022calvin, nasiriany2022learning}, which is more cost-effective to acquire but still demands a certain level of human supervision. Instead of relying on additional human data, we propose utilizing pre-defined primitives and data acquired from random primitive rollouts.
\subsection{Skill-based Imitation Learning}
Skill-based imitation learning extracts low-level temporally-extended sensorimotor behaviors as skills from expert demonstrations~\cite{zhang2018deep, Rajeswaran-RSS-18} or task-agnostic play data~\cite{lynch2020learning, mees2022calvin, nasiriany2022learning} and emulates the high-level behavior observed in the expert demonstrations to guide the execution of these low-level skills. 
A common approach involves joint learning of low-level skills and a high-level policy, with skills acquired in an unsupervised manner~\cite{shankar2019discovering, shankar2020learning, konidaris2012robot, krishnan2017ddco}. These skills can be either discrete~\cite{zhu2022bottom} or continuous in a latent space~\cite{nasiriany2022learning, pertsch2021accelerating}. Unsupervised learning obviates the need for additional human annotation but often results in skills with limited reusability and low generalization capability. Alternatively, some research focuses on learning a high-level policy and low-level skills from structured demonstrations with additional segmentation labels using supervised learning, relying on either weak~\cite{shiarlis2018taco} or strong human supervision~\cite{huang2019neural}. In our work, we use pre-built parameterized behavior primitives as low-level skills, which are highly robust, reusable, and generalizable.
Furthermore, our method requires only raw sensory demonstrations without the need for additional human annotations.

\subsection{Learning with Behavior Primitives}
One line of research focuses on policy learning with primitives, which involves augmenting the motor action space through the integration of parameterized primitives~\cite{nasiriany2022augmenting, chitnis2020efficient, hausknecht2015deep, dalal2021accelerating, lee2019learning, strudel2020learning}. Dalal et al.~\cite{dalal2021accelerating} propose to manually specify a comprehensive library of robot action primitives. These primitives are carefully parameterized with arguments that are subsequently fine-tuned and learned by an RL policy. Similarly, Nasiriany et al.~\cite{nasiriany2022augmenting} augments standard RL algorithms by incorporating a pre-defined library of behavior primitives. Chitnis et al.~\cite{chitnis2020efficient} decomposes the learning process into learning a state-independent task schema.
The discrete-continuous augmented action space imposes a significant exploration burden in RL. 
Recently, Chen et al.~\cite{chen2023predicting} proposed an imitation learning framework that integrates primitives for solving stowing tasks, which involves a complex graph construction for Graph Neural Networks to predict forward dynamics. Another tangential work by Shi et al.~\cite{shi2023waypoint} decomposes demonstrations into sequences of waypoints, which are interpolated through linear motion. The interpolated linear motion between waypoints can be regarded as a type of primitive. In contrast, our primitive-based framework can be viewed as a more versatile form of waypoint extraction, capable of encompassing a broader spectrum of skills.

\section{Method}
    We introduce \name{}, our primitive-based imitation learning framework, which decomposes complex, long-horizon tasks into concise, simple sequences of primitives. We begin by formulating the problem and providing an overview of our framework. We then describe two components of our framework: the trajectory parser and the policy.
    \subsection{Problem Formulation}
    We formulate a robot manipulation task as a Parameterized Action Markov Decision Process (PAMDP)~\cite{masson2016reinforcement}, defined by the tuple $\mathcal{M}=(\mathcal{S}, \mathcal{A}, \mathcal{P}, p_0, \mathcal{R}, \gamma)$ representing the continuous state space $\mathcal{S}$, the discrete-continuous parameterized action space $\mathcal{A}$, the transition probability $\mathcal{P}$, the initial state distribution $p_0$, the reward function $\mathcal{R}$, and the discount factor $\gamma$. In our setting, the motor action space is afforded by the primitives into discrete-continuous primitive action space, $a = (p, x),~a\in \mathcal{A},~p\in \mathcal{L}, x \in \mathcal{X}_p$, where $\mathcal{L}$ is a discrete set of primitive types and $\mathcal{X}_p$ is the parameter space of primitive $p$. We aim to learn a policy, $\pi(a|s) = \pi(p,x|s)$, $s\in \mathcal{S}$, to maximize the expected sum of discounted rewards. We assume access to a small set of task demonstrations for imitation learning.
    
    In our framework, we first parse task demonstrations into concise primitive sequences to reduce the complexity and temporal horizon of imitation learning. The challenge in parsing is to map unsegmented demonstrations into sequences of parameterized primitive actions. To establish this mapping, we build an IDM capable of identifying segments of task demonstrations into primitives. Utilizing the learned IDM, we develop a trajectory parser that uses dynamic programming to determine the optimal primitive sequences derived from task demonstrations. Subsequently, we train a policy from parsed primitive sequences to compose primitives via imitation learning. By leveraging primitives, the policy only needs to focus on primitive selection and their parameters rather than low-level motor actions. See Fig. \ref{fig:pipeline} for an overview of our framework.
    
    \subsection{Trajectory Parser}
    We develop a trajectory parser to parse task demonstrations into primitive sequences. This parser comprises an IDM and a dynamic programming algorithm. The IDM learns the probability of mapping from segments of task demonstrations to primitives. The dynamic programming algorithm determines the optimal primitive sequence by maximizing the product of probabilities in the parsed primitive sequences.
        \subsubsection{Inverse Dynamics Model}
        To parse a demonstration with primitives, it is important to identify behaviors shown in the demonstration that can be reproduced using a primitive. Toward this objective, we seek the initial and final states of behaviors and develop an IDM to infer primitives that can transition from a specified initial state to a targeted final state.
        We construct the IDM, $\text{IDM}(p, x|s, s')$, which predicts a primitive type
        $p$ and its parameters $x$ based on a pair of initial and final states $(s, s')$. The predicted primitive type falls within a categorical distribution that encompasses the types of primitives contained in the pre-built primitive set, in addition to an ``other'' category. Given that not all segments of a demonstration can be reproduced by a primitive within our pre-built primitive set, we introduce a new category named ``other''. This category is designated for classifying pairs of initial and final states that do not correspond to any of the predefined types of primitives. The predicted parameters belong to a continuous distribution, representing the parameter space associated with the primitive.
        
        To collect the training dataset for the IDM, we introduce a self-supervised data collection procedure by randomly executing primitives and atomic motor actions within the environment. Specifically, our random policy either uniformly samples a primitive type $p$ from a pre-built set or selects a random atomic motor action, each with a 50\% probability. If a primitive is sampled, its parameters $x$ are randomly chosen and the primitive is executed. Otherwise, the sampled motor action is executed directly. This process yields a set of trajectories comprising the rollouts of these primitives. These primitive rollouts are subsequently utilized to train the IDM in a supervised manner. To gather data for the ``other'' category, we randomly sample a selection of state pairs from the generated trajectories. \textcolor{black}{Specifically, we uniformly sample $K$ state pairs in each episode and label them as ``other'' category, enabling us to balance the dataset by selecting an appropriate $K$ (see Alg.~\ref{alg:rollout}). To further address data imbalance during training, we reweight the data for each primitive type $p$ in the IDM training dataset by a factor of $1 / (\text{number of rollouts for primitive type } p)$}.

        The collected dataset is an offline dataset and differs from online RL samples. The data collection policy we used is not task-specific, generating a domain-specific dataset. Consequently, all tasks within the same domain share a single IDM, requiring only one dataset collection procedure. 
                
        To enhance training data quality, we filter out unsuccessful rollouts using the success criteria for each primitive. For example, a successful grasping primitive is defined by the gripper securely holding an object. Without this filtering, many rollouts would include ineffective actions, degrading the accuracy and quality of the learned IDM and demonstration segmentation.
        
        Gathering successful primitive rollouts through uniform parameter sampling is inefficient. To reduce the sampling burden and speed up data collection, we use a prior distribution of Gaussian mixtures centered on task objects. This directs the agent to interact more effectively with objects, improving rollout success rates.
        The pseudo-code for the data generation process is summarized in Alg.~\ref{alg:rollout}.
        \setlength{\textfloatsep}{3mm}
        \begin{algorithm}
            \caption{Self-Supervised Data Collection Procedure}
            \label{alg:rollout}
            \begin{algorithmic}[1]
                \STATE \textbf{Notations}
                \STATE \hspace*{2mm} $\mathcal{D}$: training dataset of $\text{IDM}$
                \STATE \hspace*{2mm} $\mathcal{C}_p$: number of episodes during data generation
                \STATE \hspace*{2mm} $\mathcal{M}$: horizon of episodes
                \STATE \hspace*{2mm} $\mathcal{K}$: number of negative samples for each episode
                \STATE 
                \FOR{$ e \gets 1, 2, \cdots \mathcal{C}_p$}
                    \FOR{$i \gets 1, 2, \cdots\mathcal{M}$}
                        \STATE Sample a primitive and its parameters $(p^i, x_p^i)$ or\\an atomic motor action $a_{t_i}$ and set $p^i = \text{atomic}$
                        \STATE Execute it and get $\tau^i = (s_{t_i}, a_{t_i}, s_{t_i + 1}, ..., s_{t_{i+1}})$
                        \IF {$p^i \neq$ atomic \AND \texttt{is\_success}($\tau^i, p^i$)}
                            \STATE $\mathcal{D} \gets \mathcal{D} \cup \{(s_{t_i}, s_{t_{i+1}}, p^i, x_p^i)\}$
                        \ENDIF
                    \ENDFOR
                    \STATE Sequence $\{\tau^i\}_{i=1}^M$ into an episode trajectory:
                    \STATE \hspace{2mm} Get $\tau = (s_0=s_{t_0}, a_0, ..., s_{t_1}, a_{t_1}, ..., s_{t_2}, ... , s_{t_M})$
                    \FOR{$k \gets 1, 2, \cdots K$}
                        \STATE Sample a segment $\tau'=(s_{j}, a_{j}, ..., s_{l})$ from $\tau$ 
                        \STATE $\mathcal{D} \gets \mathcal{D} \cup \{(s_j, s_l, \texttt{other}, \texttt{none})\}$
                    \ENDFOR
                \ENDFOR
            \end{algorithmic}
        \end{algorithm}
        \subsubsection{Dynamic Programming Algorithm}
        The learned IDM predicts a hybrid discrete-continuous distribution when provided with a pair of input initial and final states. In this context, the probability $\text{IDM}(p,x|s, s')$ represents the likelihood of interpreting a segment between states $s$ and $s'$ as belonging to the primitive type $p$ and its parameter $x$. Once this mapping is established, the remaining task is to identify an optimal sequence of primitives that maximizes the probability of being parsed from the task demonstrations. To optimize the likelihood of primitive sequences interpreted from given task demonstrations, we leverage dynamic programming to find the optimal segmentation with a maximal product of probabilities in the parsed primitive sequences. 

        Specifically, considering a task demonstration denoted as $\tau = (s_0, a_0, ..., s_\mathcal{T})$ where $a_t$ represents low-level motor actions.
        We define the objective function in our dynamic programming process as $f(i)$, which represents the probability of decomposing $(s_0, a_0, ..., s_i)$ into an optimal sequence of primitives $\left(s_0, (p_0, x_0), s_{j_1}, (p_{1}, x_{1}), ..., s_i\right)$.
        We iteratively update the objective function
        \begin{equation}
            f(i) = \max\limits_{p,x,t<i} f(t) \cdot (\alpha \cdot \text{IDM}(p, x|s_t, s_i))
        \end{equation}
    by maximizing the product of probabilities of primitive sequences. We multiply a factor $\alpha$ to $\text{IDM}(p, x|s_t, s_i)$, where $\alpha$ is a small constant ($\alpha = 0.0001$ in our implementation) if \textcolor{black}{$p$ is in ``other'' category}; otherwise, $\alpha = 1$. We use the factor $\alpha$ to penalize mappings to the category ``other''. 
    Upon completing the dynamic programming, we can extract the optimal primitive sequences denoted as $(s_0, p_0, x_0, s_{t_1}, p_1, x_1, \ldots, s_{t_\mathcal{M}}=s_\mathcal{T})$ from the final value $f(\mathcal{T})$, where $\mathcal{M}$ represents the length of the segmented sequence.        
    \subsection{Policy Learning}
        In this section, we describe the process of learning the policy from parsed primitive sequences. We train the policy $\pi(p, x|s)$ with behavioral cloning using the segmented primitive sequences.
        
        Given that the segmented primitive sequences are notably shorter than the task demonstrations, leading to significantly smaller segmented data than the size of the demonstrations, we introduce a stepwise augmentation technique to enrich the segmented data and increase the scope of supervision for imitation learning. We assume that beginning from any point within the demonstration, the decomposition will remain consistent in subsequent segments. In other words, we presume that start and end points of each segment in parsed sequences are the same across all ``suffixes'' of demonstrations, where a suffix is defined as a portion of the demonstration beginning from any intermediate state and continuing to the final state.
        With this assumption, for each segmented primitive sequences $\tau' = (s_0, (p_0, x_0), s_{t_1}, (p_1, x_1), ..., s_\mathcal{T})$ parsed from a task demonstration $\tau = (s_0, a_0, ..., s_\mathcal{T})$, we can get an augmented tuple $(s_l, p'_d, x'_d, s_{t_{d+1}})$ at each timestep $l$, where $s_{t_d} < l < s_{t_{d+1}}$, $(p'_d, x'_d) = \operatorname*{argmax}\text{IDM}(. | s_l, s_{t_{d+1}})$. 
        We incorporate these augmented tuples $\{(s_l, p'_d, x'_d)\}$ into the training dataset of policy.

        To leverage additional prior knowledge, we pretrain the policy using the training dataset of IDM and fine-tune the policy using parsed primitive sequences.

    \subsection{Implementation Details}
    \label{sec:implement}
    We use the primitives from Nasiriany et al.~\cite{nasiriany2022augmenting} to implement our library of primitives with minor modifications. \textcolor{black}{These task-independent, hard-coded APIs can be directly adapted to new situations within the same domain, as these APIs only require robot proprioceptive information as input.} We implement the following four primitives:
    \begin{itemize}
        \item \textbf{Reaching:} The robot moves its end-effector to a target location $(x,y,z)$ and yaw angle $\psi$ in a collision-free path.
        \item \textbf{Grasping:} Same behavior and parameters as the reaching primitive, followed by the robot closing its gripper.
        \item \textbf{Placing:} Same behavior and parameters as the reaching primitive, followed by the robot opening its gripper.
        \item \textbf{Pushing:} The robot reaches a starting location $(x,y,z)$ at a yaw angle $\psi$ in a collision-free path and then moves its end-effector by a displacement $(\delta_x, \delta_y, \delta_z)$.
    \end{itemize}
    \vspace{-7pt}
    ~\\
    To reduce the complexity of the mapping, we factorize the IDM into a primitive IDM, i.e. $\text{IDM}_\text{prim}(p|s, s')$, capable of identifying primitive types and a parameter IDM, i.e. $\text{IDM}_\text{param}(x|s, s', p)$, capable of predicting parameters of the primitive, where
        \begin{equation}
        \text{IDM}(p, x|s, s')=\text{IDM}_\text{prim}(p|s, s')\cdot\text{IDM}_\text{param}(x|s, s', p).
        \end{equation}
    \textcolor{black}{Our primitive IDM and parameter IDM are two separate networks. Both networks take a pair of observations, $s$ and $s'$, and encode these observations with a pair of ResNet-18 encoders. A 2-layer MLP follows the image encoder in both networks.
    The primitive IDM outputs a Softmax distribution over primitives, while the parameter IDM outputs a Gaussian mixture model over parameter values. Our model architecture is similar to the default architecture in the RoboMimic framework~\cite{mandlekar2021matters}, as the paper reported that ResNet encoder and GMM policy uniformly perform better than other design choices.}
    Similarly to the IDM, we factorize the policy $\pi(p, x|s)$ into a primitive policy, i.e. $\pi_\text{prim}(p|s)$, and a parameter policy, i.e. $\pi_\text{param}(x|s, p)$. During deployment, the primitive policy first predicts the next primitive type $p$ given the current state $s$. Following this, the parameter policy predicts the corresponding parameters $x$ given inputs $s$ and $p$. We train these two policy networks separately with behavioral cloning. 
    These networks take a single observation as input (the current observation $s$) and have architectures and output spaces similar to their counterparts in the IDM. 
    We use RoboMimic~\cite{mandlekar2021matters} to implement and train these networks.
\section{Experiments}
    Our experiments are designed to answer the following questions: 1) How does \name{}  perform for imitation learning in low-data regimes? 2) Which design choices are critical to \name{}? 3) How effective is the trajectory parser in \name{}? 4) Is \name{} feasible for practical deployment to real-world robot tasks? 5) \textcolor{black}{Are learned skills compatible with \name{}?}

    \begin{figure}
        \centering
        \includegraphics[width=1.\linewidth]{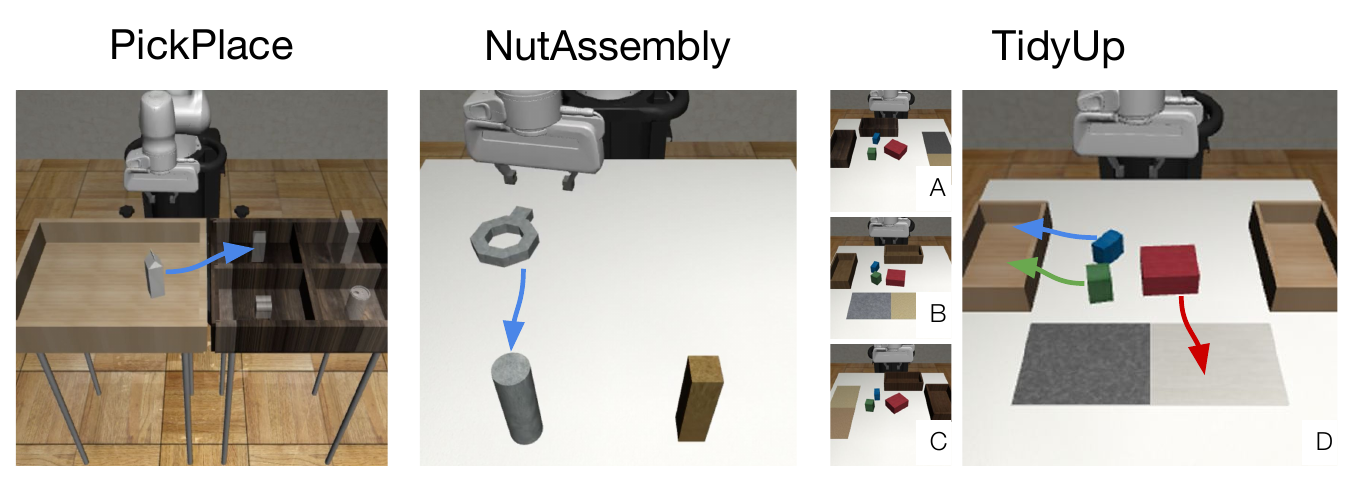}
          \caption{\textbf{Simulated Tasks}. We perform evaluations on three tasks from the RoboSuite simulator~\cite{zhu2020robosuite}. The first two, \texttt{PickPlace} and \texttt{NutAssembly}, are from the RoboSuite benchmark, with \texttt{NutAssembly} featuring less initial randomization than the original task. We introduce a third task, \texttt{TidyUp}, to study long-horizon tasks and test the inverse dynamics model's generalization to unseen environments. We create four environment variants in this domain, denoted as (A, B, C, D). \texttt{TidyUp} task is designed in environment (D), and we collect human demonstrations for \texttt{TidyUp} in the same environment (D). To gauge the inverse dynamics model's generalization capability, we train two IDMs: IDM-D, based solely on data from environment (D), and IDM-ABC, trained on data from environments (A, B, C). While IDM-D is our default model for experiments, we use IDM-ABC to evaluate generalization in unseen environments.
          }
          \vspace{1mm}
          \label{fig:sim-tasks}
    \end{figure}
        \begin{figure}
            \centering
            \vspace{5mm}
          \includegraphics[width=0.8\linewidth]{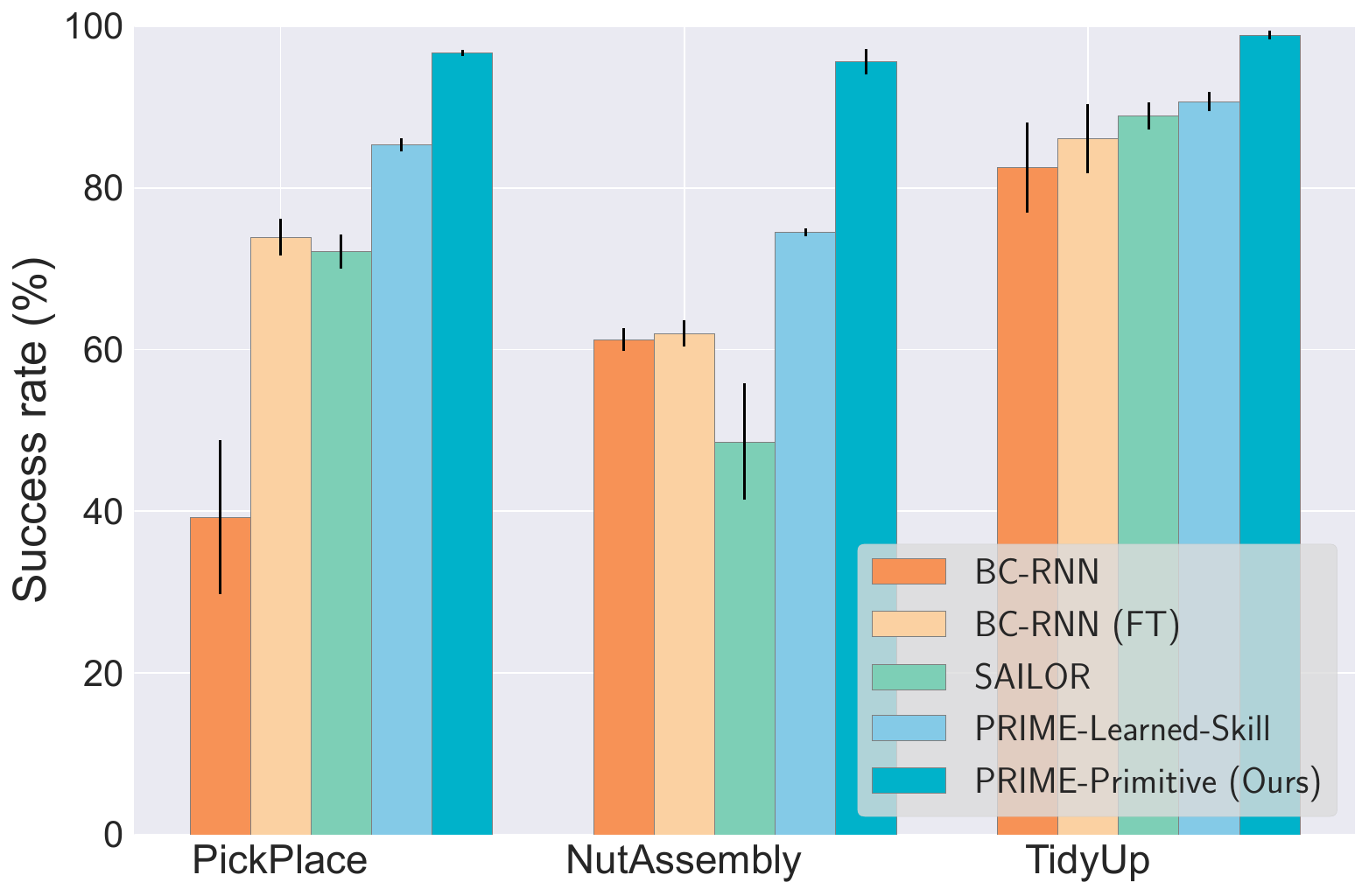}
          \caption{\textbf{Quantitative evaluation in three simulated tasks.}  Our method significantly outperforms state-of-the-art imitation learning approaches, with success rates surpassing 95\% in all three tasks.
          }
          \label{fig:main}
          \vspace{4mm}
        \end{figure}
        \begin{figure*}
            \centering
            \vspace{2mm}
              \includegraphics[width=.9\linewidth]{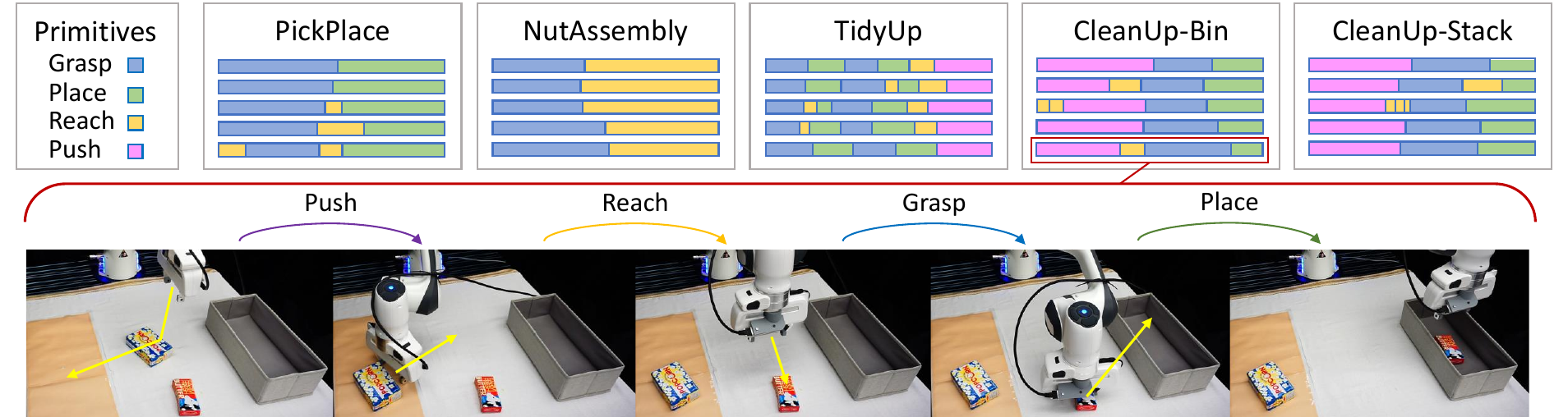} 
              \caption{\textbf{Visualization of output primitive sequences from trajectory parser.} For each task, we select five human demonstrations and visualize the segmented primitive sequences as interpreted by the trajectory parser.}
              \label{fig:vis}
            \vspace{4mm}
        \end{figure*}   
    \subsection{Experimental Setup} 
    \label{sec:exp-setup}
    \subsubsection{Manipulation Tasks}
\begin{figure}
        \vspace{1mm}
            \centering
          \includegraphics[width=1.0\linewidth]{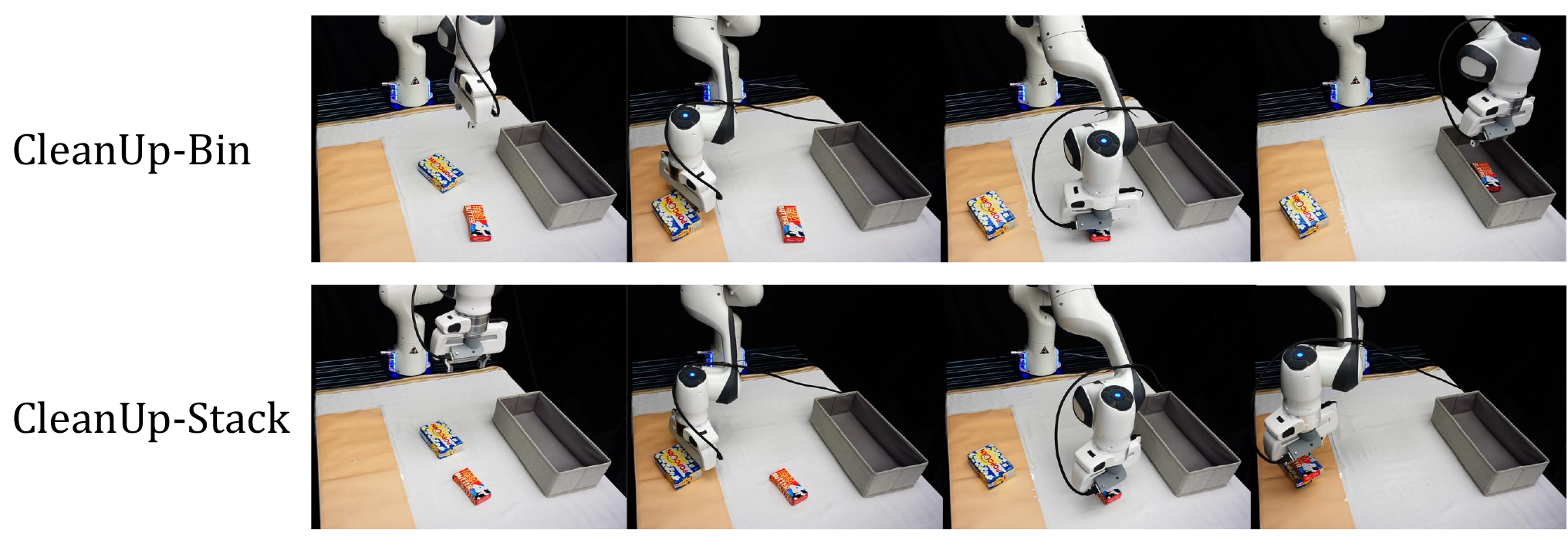}
          \caption{\textbf{Real-world Tasks.} In \texttt{CleanUp-Bin}, the robot must push the popcorn box to the serving area and place the butter inside the bin. In \texttt{CleanUp-Stack}, the robot must push the popcorn box to the serving area and stack the butter on top of the popcorn.}
          \label{fig:real-world}
          \vspace{5mm}
    \end{figure}

        We validate our approach and examine the above questions in simulated and real-world tasks. We perform evaluations in three simulated tasks from the RoboSuite simulator~\cite{zhu2020robosuite} (see Fig. \ref{fig:sim-tasks}) and two real-world tasks (see Fig. \ref{fig:real-world}):
        \begin{itemize}
            \item \textbf{PickPlace}. The robot picks up a milk carton from the left and places it in the corresponding bin.\vspace{1mm}
            \item \textbf{NutAssembly}. The robot picks up the nut and inserts it over the peg. The high precision of nut insertion is challenging when learning under a low-data regime.\vspace{1mm}
            \item \textbf{TidyUp}. A new domain to study long-horizon tasks and generalization of IDM in unseen environments. The setup includes a large box, 2 smaller boxes, 2 bins, and a table mat. Across four environment variants, objects differ in placement and texture. The robot's task is to move the large box onto the table mat and place the smaller boxes into the bins. See Fig. \ref{fig:sim-tasks} for the detailed design.\vspace{1mm}
            \item \textbf{CleanUp}. This domain replicates the real-world experimental setup introduced by Nasiriany et al.~\cite{nasiriany2022augmenting}, involving two tasks: \texttt{CleanUp-Bin} and \texttt{CleanUp-Stack}. The tasks require the robot to push a popcorn box to a serving area and move butter to a target location. See Fig.~\ref{fig:real-world} for the tasks. 
    \end{itemize}
    We employ a Franka Emika Panda robot with operational space control for all tasks. 
    In simulation and the real world, we opt for 5-DoF control at 20 Hz: 3 for end-effector position $(\delta_x, \delta_y, \delta_z)$, 1 for yaw orientation, and 1 for the gripper. The agent observes wrist-camera and third-person images and receives proprioceptive data. Vision-based policies are trained for all tasks, with only \textbf{30 human demonstrations} per task. For the training data of IDM, we collect 1M transitions for \texttt{PickPlace}, 3M transitions for \texttt{NutAssembly}, and 5M transitions for \texttt{TidyUp}. Notably, all tasks are completed based on the \textbf{same set of primitives}.  
    \subsubsection{Evaluation Protocol}
        In the simulation, we run 50 trials for each checkpoint, averaging success rates from the top 5 checkpoints per seed. The policy is executed for 20 trials per checkpoint on the real robot. We choose the best checkpoint from each seed. We compute the mean and standard deviation of success rates over the three seeds.
    \subsubsection{Baselines}
        We contrast our method with three imitation learning baselines: BC-RNN \cite{mandlekar2021matters}, its fine-tuning variant BC-RNN (FT), and SAILOR~\cite{nasiriany2022learning}. BC-RNN uses LSTM as the backbone network architecture and learns through behavioral cloning. BC-RNN (FT) pre-trains on prior data and fine-tunes with target task demonstrations.
        SAILOR pre-trains a skill encoder to extract skill latents from prior data and learns the policy using task demonstrations and pre-trained skill representations through imitation learning. The prior data for pre-training BC-RNN (FT) and SAILOR consists of all low-level transitions in the training dataset of IDM in \name{}. \textcolor{black}{\name{}-Learned-Skill runs our framework \name{} with pretrained primitives which learn from human motions by first manually segmenting the demonstrations into sequences of primitives, then training each primitive using the data from its corresponding segments. The set of primitive types remains the same as the hard-coded primitives: $\{$reach, grasp, place, push$\}$. To make a fair comparison, we use the default network architecture in RoboMimic~\cite{mandlekar2021matters} to implement BC-RNN, using ResNet-18 as the image encoder and a GMM policy, similar to our method. SAILOR utilizes a VAE architecture for skill encoding and decoding and also uses ResNet-18 and a GMM.}
    \subsection{Experimental Results}
        \begin{table}[]
            \vspace{0.7cm}
            \caption{Success rates in simulation tasks for ablation studies. }
            \label{tab:ablation_study}
            \centering
            \begin{tabular}{lccc}
            \toprule
            Task     &Ours    &No Pretraining   &Greedy Algo \\ \midrule
            \texttt{\textcolor{black}{PickPlace}} &$\textcolor{black}{\mathbf{0.967 \pm 0.004}}$     & \textcolor{black}{$0.463 \pm 0.004$}    & \textcolor{black}{$0.881 \pm 0.064$}          \\ 
            \texttt{\textcolor{black}{NutAssembly}} &$\textcolor{black}{\mathbf{0.956 \pm 0.016}}$     &\textcolor{black}{$0.705 \pm 0.005$}    & \textcolor{black}{$0.554 \pm 0.080$}          \\ 
            \texttt{TidyUp} &$\mathbf{0.989 \pm 0.005}$     & $0.944 \pm 0.003$    & $0.859 \pm 0.004$          \\ 
             \bottomrule
            \end{tabular}
            \vspace{-0.1cm}
        \end{table}
        \begin{table}[]
            \vspace{0.7cm}
            \caption{Effectiveness of trajectory parser.} 
            \label{tab:interpreter}
            \centering
            \begin{tabular}{lcc}
            \toprule
            Task                       &Success Rate                  &Primitive Seq Len / Demo Len   \\ \midrule
            \texttt{PickPlace}            & $0.967 \pm 0.027 $     & $3.5~/~314$    \\ 
            \texttt{NutAssembly}          &$0.956 \pm 0.031 $     & $2.1~/~232$ \\ 
            \texttt{TidyUp}             & $ 0.911 \pm 0.016$    & $6.6~/~403$   \\ \bottomrule
            \end{tabular}
            \vspace{-0.1cm}
        \end{table}
        \begin{table}[]
            \vspace{6mm}
            \caption{Generalization capability of IDM.}
            \label{tab:idm}
            \centering
            \begin{tabular}{lccc}
            \toprule
            IDM                  & Ours                  & BC-RNN        & BC-RNN (FT) \\ \midrule
            \texttt{IDM-D}                     & $ \mathbf{0.989 \pm 0.005}$     & $ 0.825 \pm 0.056 $    & $ 0.861 \pm 0.043$             \\ 
            \texttt{IDM-ABC}  & $ \mathbf{0.975 \pm 0.030}$     & $ 0.825 \pm 0.056 $    & $ 0.860 \pm 0.042$             \\ 
            \bottomrule
            \end{tabular}
            \vspace{-0.1cm}
        \end{table} 
        \subsubsection{Quantitative Results}
        \label{sec:exp_main}
        Fig. \ref{fig:main} demonstrates our method's substantial superiority over all baselines, achieving success rates exceeding 95\% across all tasks with remarkable robustness. This showcases the effectiveness of our approach in achieving data-efficient imitation learning through the decomposition of raw sensory demonstrations into concise primitive sequences. Furthermore, our policy often attempts the same primitive type and slightly adjusts primitive parameters after a failure. The results indicate that our primitive policy is more capable of learning this recovery attempt following a failure than baseline policies which require predicting a sequence of actions for recovery. \textcolor{black}{Failures in our method are typically caused by prediction errors in the parameter policy.}

        \textcolor{black}{\name{}-Learned-Skill outperforms all other baselines, confirming that learned skills are compatible with our \name{} framework. However, the fact that \name{}-Learned-Skill performs worse than \name{} with hard-coded primitives suggests that primitives pretrained from a small set of annotated human demonstrations are suboptimal, leading to a suboptimal low-level controller.}
        
        \subsubsection{Ablation Studies}
        We perform ablation studies in all three simulation tasks to study the importance of design choices in our framework. We compare our methods with ablations: 1) without policy pretraining (\textbf{No Pretraining}), 2) utilizing a greedy algorithm to interpret demonstrations into primitive sequences instead of dynamic programming (\textbf{Greedy Algorithm}), where the greedy algorithm selects next primitive by examining all states in the demonstration that come after the current state and choosing the one with the highest probability.
        As shown in Table \ref{tab:ablation_study}, omitting policy pre-training or substituting dynamic programming with the greedy algorithm leads to decreased performance, highlighting their essential roles. 
        \subsubsection{Model Analysis}
        \textbf{Effectiveness of the Trajectory Parser.}
        To evaluate our trajectory parser's performance, we execute the parsed primitive sequences in the simulated task environment. As shown in Table \ref{tab:interpreter}, our trajectory parser consistently achieves over 90\% success rates. Moreover, the ratio of average primitive sequence length (primitive seq len) to average demonstration length (demo len) illustrates that the trajectory parser is capable of reducing the task horizon from hundreds of steps to just a few steps. In Fig. \ref{fig:vis}, we present a visualization of the output primitive sequences generated by the trajectory parser.
        
        \textbf{Generalization Capability of the IDM.} 
        We assess the generalization capability of IDM in \texttt{TidyUp} task. As depicted in the caption of Fig. \ref{fig:sim-tasks}, we train an IDM-ABC on data collected from the environment (A, B, C) and an IDM-D on data collected merely from the environment (D). As shown in Table \ref{tab:idm}, the policy performance achieved using IDM-ABC is comparable to that achieved using IDM-D, highlighting the IDM's generalization capability in an unseen environment.
    \subsection{Real-World Evaluation}
    \label{sec:real-world}   
We evaluate the performance of \name{} against an imitation learning baseline (BC-RNN) on two real-world \texttt{CleanUp} task variants: \texttt{CleanUp-Bin} and \texttt{CleanUp-Stack}. 
To ensure safe data collection, we perform self-supervised data collection in simulation to train an IDM and apply it to real-world demonstrations. The real-world observations include camera images, object poses, and robot proprioceptive states. Our state-based IDM uses object poses and proprioceptive states, transferring directly to real-world demonstrations. A single IDM segments demonstrations for two tasks within the same domain. Additionally, we develop a visual-based policy to predict motor actions from camera images and robot states. For object poses on the real robot, we use a pose estimator~\cite{tremblay2018dope}.

The results in Table~\ref{tab:CleanUp} show our method's notable advantage over BC-RNN, primarily due to its effective task scaffolding. Similar to the results in the simulation, our method also exhibits robust recovery behavior on real robots. 
Our method has a success rate below 70\% in \texttt{CleanUp-Stack}, mainly because stacking demands greater precision and sim2real training for IDM leads to errors.

        \begin{table}[]
            \centering
            \vspace{1.cm}
            \caption{Results on the real robot.}
            \label{tab:CleanUp}
            \begin{tabular}{lcc}
            \toprule
            Task                        & Ours       & BC-RNN  \\ \midrule
            \texttt{CleanUp-Bin}        & $\mathbf{0.900 \pm 0.041}$    & $0.417 \pm 0.246$              \\ 
            \texttt{CleanUp-Stack}      & $\mathbf{0.683 \pm 0.062}$    & $0.483 \pm 0.131$              \\ \bottomrule
            \end{tabular}            
            \vspace{0.1cm}
        \end{table}

\vspace{0mm}
\section{Conclusion}
We present \name{}, a data-efficient imitation learning approach that decomposes task demonstrations into sequences of primitives and leverages imitation learning to acquire the high-level control policy for sequencing parameterized primitives. 
\textcolor{black}{While we have already evaluated \name{} with pretrained primitives from annotated demonstrations, a promising direction for future research is to learn a scalable library of low-level skills and compose these diverse skills. These skills can include hard-coded primitives, learned primitives, and skills pretrained from large datasets.} This approach holds the potential to facilitate curriculum learning, enabling the progressive acquisition of increasingly complex tasks. 
A limitation of this study is the use of sim2real experiments for IDM training, which may not be fully applicable to challenging real-world tasks. Extending IDM training to real-world settings is left for future research. \textcolor{black}{Another limitation is that all tasks in this work can be fully decomposed into primitives from the primitive library. Extending our work to include tasks that are not fully decomposable would enhance the generalizability of our framework.}
\section*{ACKNOWLEDGMENT}
 The authors would like to thank Yifeng Zhu, Jake Grigsby, Mingyo Seo, Rutav Shah, and Zhenyu Jiang for their valuable feedback. Tian Gao's visit to UT Austin was supported by IIIS, Tsinghua University. Quantao Yang's visit to UT Austin was supported by his advisor, Todor Stoyanov, and the Wallenberg AI, Autonomous Systems, and Software Program (WASP).
 This work has been partially supported by the National Science Foundation (EFRI-2318065, FRR-2145283), the Office of Naval Research (N00014-22-1-2204), UT Good Systems, and the Machine Learning Laboratory.


\vspace{-1mm}
\printbibliography

\end{document}